\documentclass[10pt,twocolumn,letterpaper]{article}

\usepackage[pagenumbers]{cvpr} %

\usepackage{graphicx}
\usepackage{amsmath}
\usepackage{amssymb}
\usepackage{booktabs}
\usepackage{multirow}
\usepackage{adjustbox}

\usepackage[utf8]{inputenc} %
\usepackage[T1]{fontenc}    %
\usepackage{url}            %
\usepackage{booktabs}       %
\usepackage{amsfonts}       %
\usepackage{nicefrac}       %
\usepackage{microtype}      %
\usepackage{xcolor}         %
\usepackage{xspace}
\usepackage{mathtools}

\usepackage{bbm}

\usepackage[pagebackref,breaklinks,colorlinks]{hyperref}

\usepackage[capitalize]{cleveref}
\crefname{section}{Sec.}{Secs.}
\Crefname{section}{Section}{Sections}
\Crefname{table}{Table}{Tables}
\crefname{table}{Tab.}{Tabs.}

\makeatletter
\DeclareRobustCommand\onedot{\futurelet\@let@token\@onedot}
\def\@onedot{\ifx\@let@token.\else.\null\fi\xspace}

\def\eg{\emph{e.g}\onedot} 
\def\ie{\emph{i.e}\onedot} 
 
\def\etc{\emph{etc}\onedot}

\makeatother

\newcommand{\system}{A2Summ\xspace}

\def\cvprPaperID{2943} %
\def\confName{CVPR}
\def\confYear{2023}

\begin{document}

\title{Align and Attend: Multimodal Summarization with Dual Contrastive Losses}

\author{Bo He$^{1}$\thanks{Part of this work was done when Bo was an intern at Adobe Research.}, Jun Wang$^{1}$, Jielin Qiu$^{2}$, Trung Bui$^{3}$, Abhinav Shrivastava$^{1}$, Zhaowen Wang$^{3}$\\[0.5em]
$^{1}$University of Maryland, College Park \quad\quad
$^{2}$Carnegie Mellon University \quad\quad
$^{3}$Adobe Research\\
{\tt\small \{bohe,abhinav\}@cs.umd.edu, junwong@umd.edu, jielinq@andrew.cmu.edu, \{bui,zhawang\}@adobe.com}
}
\maketitle

\begin{abstract}
The goal of multimodal summarization is to extract the most important information from different modalities to form summaries.
Unlike unimodal summarization, the multimodal summarization task explicitly leverages cross-modal information to help generate more reliable and high-quality summaries. 
However, existing methods fail to leverage the temporal correspondence between different modalities and ignore the intrinsic correlation between different samples.
To address this issue, we introduce \textbf{A}lign and \textbf{A}ttend Multimodal \textbf{Summ}arization (\system), a unified multimodal transformer-based model which can effectively align and attend the multimodal input. 
In addition, we propose two novel contrastive losses to model both inter-sample and intra-sample correlations.
Extensive experiments on two standard video summarization datasets (TVSum and SumMe) and two multimodal summarization datasets (Daily Mail and CNN) demonstrate the superiority of \system, achieving state-of-the-art performances on all datasets.
Moreover, we collected a large-scale multimodal summarization dataset BLiSS, which contains livestream videos and transcribed texts with annotated summaries.
Our code and dataset are publicly available at
~\url{https://boheumd.github.io/A2Summ/}.
\end{abstract}
\section{Introduction}

With the development in multimodal learning, multimodal summarization has drawn increasing attention~\cite{zhu2018msmo,narasimhan2021clip,khullar2020mast,chen2018abstractive,hori2019end,fu2021mm,Qiu2022MHMSMH,Qiu2022SemanticsConsistentCS,fu2022doc2ppt}.
Different from traditional unimodal summarization tasks, such as video summarization~\cite{zhang2016video,zhou2018deep,rochan2018video,fu2019attentive,zhu2020dsnet,park2020sumgraph,zhao2021reconstructive,jiang2022joint} and text summarization~\cite{nallapati2016abstractive,cheng2016neural,nallapati2017summarunner,miller2019leveraging,zhong2020extractive}, multimodal summarization aims at generating summaries by utilizing the information from different modalities.  
With the explosive growing amount of online content (\eg, news, livestreams, vlogs, \etc), multimodal summarization can be applied in many real-world applications.
It provides summarized information to the users, which is especially useful for redundant long videos such as livestream and product review videos.

\begin{figure}[t]
\centering
    \vspace{-0.15in}
    \adjincludegraphics[width=\linewidth, trim={{0.0\width} {0.0\height} {0.0\width} {0.0\height}},clip]{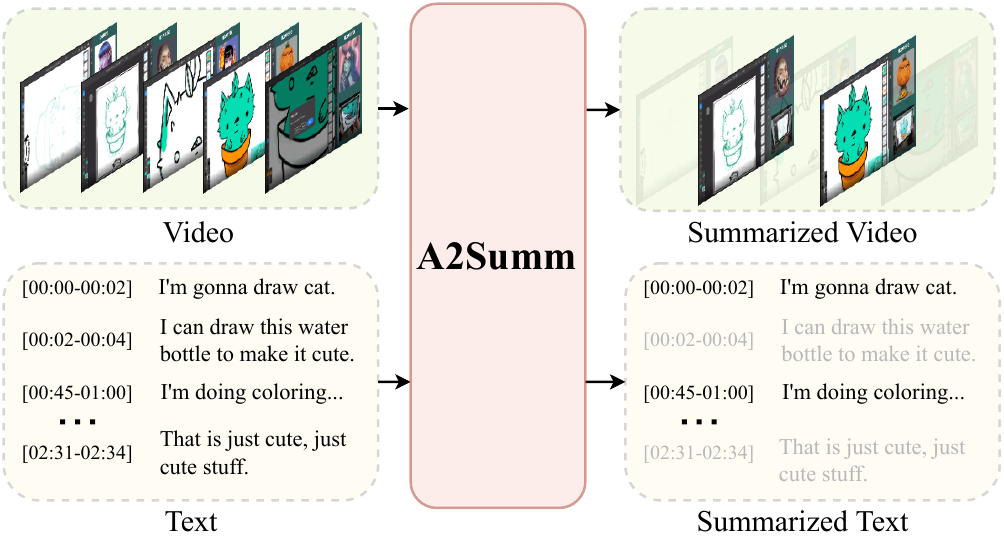}
    \caption{\system is a unified multimodal summarization framework, which aligns and attends multimodality inputs while leveraging time correspondence (\eg, video and transcript) and outputs the selected important frames and sentences as summaries. }
    \label{fig:teaser}
\vspace{-0.2in}
\end{figure}

Previous multimodal summarization methods~\cite{li2017multi,li2018multi,narasimhan2021clip,chen2018abstractive} leverage the additional modality information but can only generate the main modality summary, \ie, either a video summary or a text summary, severely limiting the use of 
complementary benefits in the additional modality. 
Recently, multimodal summarization with multimodal output (MSMO) has been explored in several studies~\cite{zhu2018msmo,li2020vmsmo,fu2020multi,fu2021mm}, which aim at generating both video and text summaries using a joint model.
Compared to previous methods, which only produce a unimodal summary, MSMO provides a better user experience with an easier and faster way to get useful information. However, we find that the existing MSMO methods still have the following limitations.
First, even if both modalities are learned together, the correspondence between different modalities is not exploited. For example, given a video and its transcripts, which are automatically matched along the time axis, no existing method utilizes the mutual temporal alignment information and treats the two modalities separately.
Second, previous works adopt simple strategies to model the cross-modal correlation by sequence modeling and attention operation~\cite{zhu2018msmo,chen2018abstractive,li2020vmsmo,fu2020multi,fu2020multi,li2020vmsmo}, which requires a large number of annotated multimodal data which is hard to obtain.

Motivated by the above observations, we propose a novel architecture for multimodal summarization based on a unified transformer model, as shown in Figure~\ref{fig:teaser}. 
First, to leverage the alignment information between different modalities, we propose alignment-guided self-attention module to align the temporal correspondence between video and text modalities and fuse cross-modal information in a unified manner.
Second, inspired by the success of self-supervised training~\cite{Arandjelovic_2017_ICCV,asano2020labelling,Patrick_2021_ICCV}, which utilizes the intrinsic cross-modality correlation within the same video and between different videos,
we propose dual contrastive losses with the combination of an inter-sample and an intra-sample contrastive loss, to model the cross-modal correlation at different granularities. 
Specifically, the inter-sample contrastive loss is applied across different sample pairs within a batch, which leverages the intrinsic correlation between each video-text pair and contrasts them against remaining unmatched samples to provide more training supervision. 
Meanwhile, the intra-sample contrastive loss operates within each sample pair, which exploits the mutual similarities between ground-truth video and text summaries and contrasts the positive features against hard-negative features.

To facilitate the research of long video summarization with multimodal information, we also collected a large-scale livestream video dataset from the web.
Livestream broadcasting is growing rapidly, and the summarization of livestream videos is still an unexplored area with great potential.
Previous video summarization datasets consist of short videos with great variations in scene transitions. On the contrary, livestream videos are significantly longer (in hours as opposed to minutes) and the video content changes much more slowly over time, which makes it even harder for the summarization task.
Besides, there has been a lack of annotated datasets with focus on transcript summarization, which can be a great complement to the livestream video summarization.
Therefore, we collect a large-scale multimodal summarization dataset with livestream videos and transcripts, which are both annotated with ground-truth summaries by selecting important frames and sentences.

To summarize, our contributions include:
\begin{itemize}
    \vspace{-0.5em}
    \item We propose \system, a unified transformer-based architecture for multimodal summarization. It can handle multimodal input with time correspondences which previous work neglects.
    \vspace{-0.5em}
    \item We present dual contrastive losses that account for modeling cross-modal information at different levels. Extensive experiments on multiple datasets demonstrate the effectiveness and superiority of our design. 
    \vspace{-0.5em}
    \item A large-scale \textbf{B}ehance \textbf{Li}ve\textbf{S}tream \textbf{S}ummarization (BLiSS) dataset is collected containing livestream videos and transcripts with multimodal summaries.
\end{itemize}

\section{Related Work}

\paragraph{Video Summarization.} 
Current techniques for video summarization can be divided into two categories, unsupervised and supervised. 
Unsupervised learning approaches, including~\cite{de2011vsumm,lee2012discovering, elhamifar2015dissimilarity, jung2019discriminative, lu2014bag, yuan2019cycle, qiu2023liveseg,zhao2014quasi, he2019unsupervised, jung2019discriminative, zhou2018deep,jung2020global} utilize different hand-crafted features to score and select the video frames without the human-annotated summaries. 
DR-DSN~\cite{zhou2018deep} explores an unsupervised reward function to tackle video summarization. 
GLRPE~\cite{jung2020global} attempts to apply self-attention with relative position representation for unsupervised video summarization. 
With the help of the annotated video summarization datasets~\cite{song2015tvsum,gygli2014creating}, numerous supervised learning methods~\cite{gong2014diverse, ji2019video, sharghi2017query, jiang2022joint, zhao2018hsa, zhu2020dsnet,zhao2021reconstructive,jiang2022joint} have been proposed in recent years to summarize videos. Among them, 
DSNet~\cite{zhu2020dsnet} formulates supervised video summarization as a temporal interest detection process. 
RSGN~\cite{zhao2021reconstructive} utilizes LSTM and GCN to model frame-level and shot-level dependencies.
iPTNet~\cite{jiang2022joint} jointly trains the video summarization task and correlated moment localization task to utilize additional moment localization data samples to boost the video summarization performance. 

\vspace{-4mm}
\paragraph{Text Summarization.}
In general, text summarization can be categorized into two groups: \emph{(i)} Extractive summarization~\cite{cheng2016neural,nallapati2017summarunner, miller2019leveraging, wang22n_interspeech, zhong2020extractive} generates output summary by identifying the salient parts of the input document. NN-SE~\cite{cheng2016neural} develops a neural attention model to select sentences or words of the input document as the output summary. SummaRuNNer~\cite{nallapati2017summarunner} employs RNN for extractive summarization. Miller~\cite{miller2019leveraging} adopts clustering algorithm in the feature space to select important sentences. 
\emph{(ii)} Abstractive summarization~\cite{nallapati2016abstractive,see2017get,paulus2017deep, zhang2020pegasus, liu2019text} performs the summarization by paraphrasing the important parts of the input document. Lead3~\cite{nallapati2016abstractive} applies the attentional encoder-decoder RNN for the task of abstractive text summarization. However, those approaches are designed for pure unimodal summarization that doesn't consider cross-modal alignment and fusion.
Recently, StreamHover~\cite{cho2021streamhover} presents an unsupervised model for transcript summarization and collects a livestream transcript summarization dataset.
Inspired by it, we collect a new livestream dataset with a much larger scale and richer multimodal annotations for the multimodal summarization task.

\begin{figure*}[t!]
    \centering
    \vspace{-0.15in}
    \includegraphics[width=0.97\textwidth]{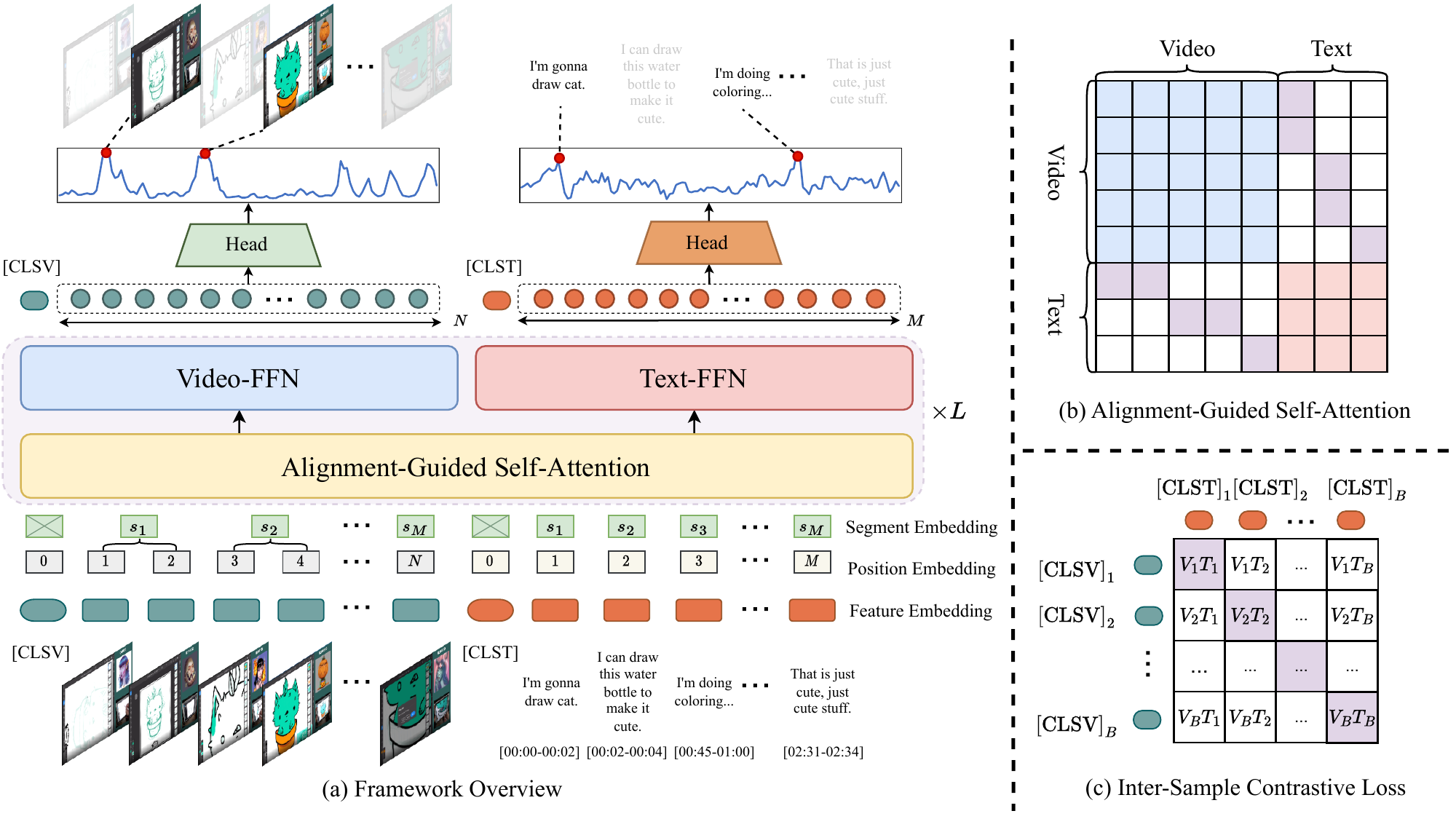}
    \vspace{-0.05in}
    \caption{(a) The overview of \system framework. Given $N$ video frames and $M$ text sequences as input, \system predicts the important frames and sentences as multimodal summaries. 
    (b) Alignment-guided self-attention module is applied to align and fuse each video and text pair. 
    (c) Inter-sample contrastive loss is calculated by maximizing the similarity of [CLSV] and [CLST] tokens from the same pair while minimizing the similarity of tokens from different pairs. $B$ is the batch size. Best viewed in color.}
    \label{fig:model}
    \vspace{-0.15in}
\end{figure*}

\vspace{-4mm}
\paragraph{Multimodal Summarization.}
Existing work~\cite{li2017multi, li2018multi, chen2018abstractive, truong2019vistanet, narasimhan2021clip} commonly utilize additional complementary modalities to enhance the feature representation for the primary modality, however, they typically generate summaries from a single modality.
For example, CLIP-It~\cite{narasimhan2021clip} builds a language-guided framework to obtain a summary video conditioned on the text. MMS~\cite{li2017multi} learns joint representations of text and images and outputs text summaries.
Recently, multimodal summarization with multimodal output (MSMO) has been explored in several studies.
Zhu et al.~\cite{zhu2018msmo} propose the first MSMO model and collect a multimodal summarization dataset with text and image modalities. 
Li et al.~\cite{li2020vmsmo} extend it with video-based news articles and adopt conditional self-attention for text and video fusion.
Recently, Fu et al.~\cite{fu2021mm} collect a multimodal dataset with more modalities included such as audio and transcript.

\section{Method}

\subsection{Overview}
Given the untrimmed multimodality input (\eg video, text, and sound), the multimodal summarization task aims at selecting the most important parts from each modality. 
Figure~\ref{fig:model}(a) illustrates an overview of our \system framework. 
The input to our model is the multi-modality (\eg, video and transcribed text in our case) with $N$ video frames and $M$ sentences.
Since each transcribed sentence has its start time and end time, the video and text modalities can be automatically aligned by the corresponding timestep.
The overall architecture can be divided into three parts: the input embedding (Sec.~\ref{sec:input}), the multimodal alignment and fusion (Sec.~\ref{sec:fusion}), and the loss function (Sec.~\ref{sec:loss}).

\subsection{Input Embedding}
\label{sec:input}
Similar to previous work~\cite{gygli2015video,zhang2016video,zhu2020dsnet,jiang2022joint}, we use pre-trained feature extraction models (\eg, GoogleNet~\cite{szegedy2015going} and RoBERTa~\cite{liu2019roberta}) to extract deep neural features for each frame and sentence. 
After feature extraction, features from different modalities are projected into a common $C$-dimensional embedding space by a linear fully connected (FC) layer.
Specifically, we denote the generated video and text features as $F\in\mathbf{\mathbb{R}^{N\times C}}$ and $S\in\mathbf{\mathbb{R}^{M\times C}}$, respectively.

For each modality, there is a special token ``[CLS]'' prepended at the start of the feature sequences, which enables a holistic representation.
Following BERT~\cite{devlin2018bert}, we add a learnable position embedding to each feature sequence so that the order information can be incorporated.
To utilize the time correspondence information between the video frames and text sentences, we add an additional learnable segment-based positional embedding at the input stage.
More precisely, each sentence has its own timestep information denoted as $[t_s, t_e]$, where $t_s$ and $t_e$ denote the start and the end time index of each sentence. 
We note that a single text sentence usually corresponds to several video frames, making $M \leq N$. 
For all frames inside each time index window $\{F_i\}_{i\in[t_s, t_e]}$, the segment embedding is shared across these frames and the corresponding sentence.
After adding these positional embeddings, the input sequences to the multimodal transformer from both modalities are concatenated along the time axis, denoted as $X \in \mathbb{R}^{(M+N)\times C}$.

\subsection{Multimodal Alignment and Fusion}
\label{sec:fusion}
\paragraph{Alignment-Guided Self-Attention.}
A core component of \system is the alignment-guided self-attention module which allows us to exploit the time correspondence between video and text modalities. 
Inspired by the superior advantages of Transformers~\cite{vaswani2017attention} in modeling different modalities (\eg, visual, language, and audio) on various multimodal tasks (\eg, visual question answering~\cite{lu2019vilbert,tan2019lxmert,wang2022image,wang2021vlmo,Wang_2022_BMVC}, vision-language pre-training~\cite{zhou2020unified,lu2016hierarchical,hu2020iterative}), we adopt the transformer architecture to align and fuse our multimodal input.
However, for the multimodal summarization task, the inputs are often untrimmed videos and text sentences, which are dominated by irrelevant backgrounds.
Directly applying global self-attention across inputs from all modalities may introduce extra noise to the multimodal fusion process.

Motivated by this observation, we propose the alignment-guided self-attention module to fuse the input across different modalities.
We formulate this process by using a masked self-attention operation in Figure~\ref{fig:model}(b). 
Specifically, an attention mask $A\in \mathbb{R}^{(N+M)\times (N+M)}$ initialized with all 0 is defined to indicate the timestep alignment information, where $N$ and $M$ denote the length of the video and text feature sequences, respectively.
For the intra-modality modeling, we follow the standard procedure with global attention operation, where features from the same modality can attend to each other such that all entries corresponding to intra-modality attention are filled with value 1 in the attention mask.
For the cross-modality attention between video and text input, we only fill in the entries from the same segment with value 1.
For example, suppose the $k^{th}$ sentence $S_k$ corresponding to the time index window $[t_s, t_e]$. We consider the frames which also lie into the same time window $[t_s, t_e]$ to be the same segment, denoted as $\{F_i\}_{i\in[t_s, t_e]}$. Then, we assign the elements of attention mask as follows $A[N+k, t_s: t_e] = 1$.
The attention mask is then applied to the attention matrix computed by the standard self-attention approach~\cite{vaswani2017attention,wang2018non,he2022asm,he2020gta}:
\begin{align}
    Q&=XW_Q,\  K=XW_K,\  V = XW_V, \label{eq:attention1} \\
    D_{i,j} &= \dfrac{A_{i,j} \text{exp}(Q_iK_j^T/\sqrt{D})}{{\sum_k A_{i,k} \text{exp}(Q_iK_k^T/\sqrt{D})}}, \label{eq:attention2}\\
    Z&=X+DV W_O, \label{eq:attention3}
\end{align}
where $i,j\in[1, M+N]$ are the entry indices of the matrix, $X$ is the concatenated input from video and text modalities, and $W_Q,W_K,W_V,W_O\in \mathbb{R}^{C\times C}$ are the linear projection matrices for generating the query, key, value, and the output. Multi-head attention~\cite{vaswani2017attention} is also adopted to improve the capacity of the attention module.
In this way, we explicitly utilize the alignment correspondence between different modalities, avoiding the negative impacts caused by noisy background frames or irrelevant sentences.

\vspace{-4mm}
\paragraph{Mixture-of-Modality-Experts.} Based on the mixture-of-modality-experts transformer~\cite{wang2021vlmo} in the multimodal tasks, after the self-attention layer, we introduce two different experts to jointly model features from different modalities including the video expert (Video-FFN), and text expert (Text-FFN) rather than the standard shared FFN \cite{vaswani2017attention}. 

\vspace{-4mm}
\paragraph{Score Prediction.}
Finally, on top of the cascaded transformer blocks, two separate score prediction branches assign relevance scores to each frame and each sentence.
Based on predicted scores, two different procedures are followed to generate the final summary.
For the standard video summarization datasets (\eg, SumMe~\cite{gygli2014creating} and TVSum~\cite{song2015tvsum}), based on the pre-processed KTS~\cite{potapov2014category} segmentation results, segment-level scores are computed from frame-level scores, and the final video summary is generated by selecting top 15\% of video durations by Knapsack algorithm.
For the multimodal summarization datasets (\eg, Daily Mail~\cite{fu2021mm}), the frames and sentences with the top highest scores are selected to generate the final summary prediction for the video and text modalities separately.

\subsection{Loss Function}
\label{sec:loss}
We employ three different loss functions to train our model, including the classification loss and the novel dual contrastive losses, which consist of the inter-sample contrastive loss and the intra-sample contrastive loss.

\vspace{-2mm}
\paragraph{Classification Loss.}
We apply the focal loss~\cite{lin2017focal} for the importance score classification, which handles the class imbalance issue by down-weighting losses for well-classified samples.
The details are shown below:
\begin{equation}
    \mathcal{L}_{\text{cls}_\text{m}}=-\frac{1}{N}\sum_{i=1}^N \left\{
    \begin{array}{ll}
        -\alpha(1{-}p_i)^\gamma \log(p_i), &\text{ if } y_i{=}1 \\
        -(1{-}\alpha) p_i^\gamma \log(1-{p_i}), &\text{ if } y_i{=}0
    \end{array}
    \right. \\
\end{equation}
\begin{equation}
    \mathcal{L}_{\text{cls}} = \mathcal{L}_{\text{cls}_\text{video}} + \mathcal{L}_{\text{cls}_\text{text}}
\end{equation}
where m could be either video or text, and $p_i$ is the predicted score for each frame/sentence while $y_i$ is the ground-truth label. If $y_i{=}1$, it indicates the $i^{th}$ frame/sentence is the key-frame/key-sentence. 
The final classification loss is the sum of the two single modality losses.

\begin{figure*}[t!]
    \centering
    \vspace{-0.2in}
    \includegraphics[width=0.97\textwidth]{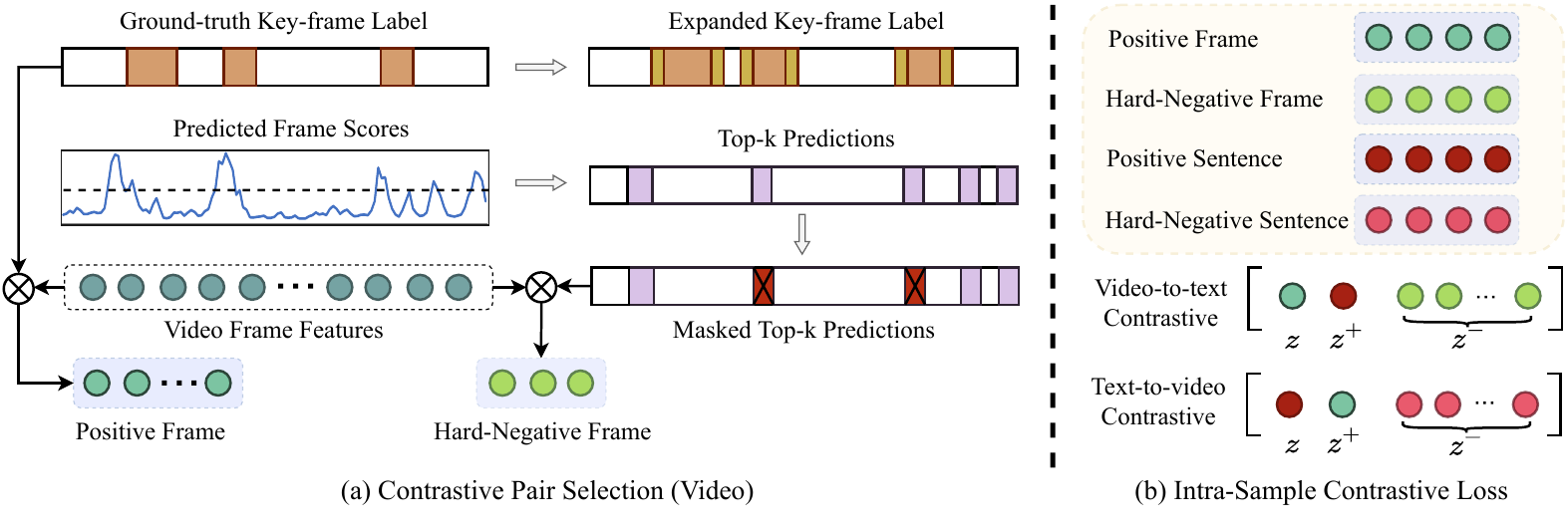}
    \vspace{-0.05in}
    \caption{(a) Contrastive pair selection process for selecting positive and hard-negative video frame features. The same procedure is applied to the text modality. The crossed red boxes denote the top predicted time steps masked out by the expanded key-frame label. (b) Intra-sample contrastive loss is applied between the selected video and text pairs. Best viewed in color.}
    \label{fig:intra_contrastive}
    \vspace{-0.15in}
\end{figure*}

\vspace{-2mm}
\paragraph{Inter-Sample Contrastive Loss.}
Driven by the success of contrastive learning in the image-language pre-training tasks~\cite{radford2021learning,he2020momentum,chen2021empirical}, we want to utilize the intrinsic relationships between each input video and text pair.
As shown in Figure~\ref{fig:model}(c), given a batch of $B$ sample pairs, we design an auxiliary inter-sample contrastive loss to predict which of the $B^2$ possible video-text pairs across a batch correctly matches and belongs to the same sample.
Specifically, we use the pre-pended [CLS] token as a holistic representation for each video and text sample.
Similar to CLIP~\cite{radford2021learning}, we maximize the cosine similarity of the video embedding [CLSV] and the text embedding [CLST] from $B$ real pairs in the batch while minimizing the cosine similarity of embeddings from the $B^2 - B$ incorrect pairs. 
Specifically, the inter-sample contrastive loss is calculated as 
\begin{align}
\begin{split}
    \mathcal{L}_{\text{inter}} = \ & \displaystyle \mathbb{E}_{z\sim \text{[CLSV]}_j, z^{+}\sim \text{[CLST]}_{j}, z^-\sim \mathcal{I}_{k\neq j}\text{[CLST]}_k}  \ell(z,z^+,z^-) \\
    + \ & \displaystyle \mathbb{E}_{z\sim \text{[CLST]}_j, z^{+}\sim \text{[CLSV]}_{j}, z^-\sim \mathcal{I}_{{k\neq j}}\text{[CLSV]}_{k} } \ell(z,z^+,z^-)
\end{split}
\end{align}
where $\ell(z,z^+,z^-)$ is the standard contrastive loss~\cite{he2020momentum} with the following equation:
\begin{multline}
    \ell(z,z^+,z^-) \\
    = \ - \log \left( \frac{\exp{(z^T \cdot z^+/\tau)}} {\exp{(z^T \cdot z^+/\tau)} + \sum_{k} \exp{(z^T \cdot z^-_k/\tau)}} \right) \label{eq:contrastive}
\end{multline}
and $\tau$ is a learnable temperature parameter.

\vspace{-4mm}
\paragraph{Intra-Sample Contrastive Loss.}
While the above inter-sample contrastive loss only considers the relationship across different samples, however, for the summarization task, to correctly detect the key-frames and key-sentences from each untrimmed video and text input, 
more fine-grained information modeling, in particular, is crucial.
It would require the model to accurately distinguish the key-frames and key-sentences from the background frames and less-related sentences.
Intuitively, the human-annotated key-frames and key-sentences share mutual information with each other. Meanwhile, they both should reveal the most salient parts from the original untrimmed video and text sequences.
For instance, for a cooking recipe video with transcribed text, the annotated key-frames and key-sentences should clearly reveal the instructions for each step. 
More importantly, these key-frames and key-sentences should be deeply correlated with each other and share similar high-level semantic meanings.
Motivated by this observation, we propose the intra-sample contrastive loss which is calculated within each video and text pair sample rather than across different sample pairs.

Specifically, we assign features associated with the pre-defined ground-truth key timesteps as positive pairs for both modalities.
To form the contrastive pairs, as pointed out by~\cite{robinson2020contrastive,kalantidis2020hard}, the quality of the negative samples is of vital importance for the effectiveness of contrastive learning. 
Therefore, we need to select the hard negative samples for video and text separately.
Specifically, since the pre-annotated non-key timesteps are negative samples, based on the prediction scores for each frame (${p_i}_{i=1}^{N}$) and sentence (${q_i}_{i=1}^{M}$), we argue that the wrongly classified timesteps with highest prediction scores are hard-negative samples. 
Intuitively, for a long untrimmed video, due to the time dependencies, the frames adjacent to the key-frames have very similar visual contents and should also be treated as the key-frames. 
However, if these frames are selected as the hard-negative samples, it tends to confuse the model and may hurt the final performance.
Therefore, we exclude those timesteps before selecting the hard-negative samples.

As shown in Figure~\ref{fig:intra_contrastive}(a), given the ground truth (GT) key-frame label, we first expand the key-frame segments on both sides to include more adjacent frames as dummy key-frames.
Then, based on the predicted scores for each timestep, we select timesteps with top-$k$ highest scores but not in the expanded GT key-frame labels as hard-negative samples. 
Here, $k_{\text{video}}=\lfloor \frac{N}{r} \rfloor$, $k_{\text{text}}=\lfloor \frac{M}{r} \rfloor$,  $r$ is a hyper-parameter controlling the total number of selected hard-negative samples.
In this way, we form contrastive pairs for both video and text modalities.
Formally, we denote the positive frames, hard-negative frames, positive sentences, and hard-negative sentences as $\mathcal{I}_{\text{PF}}$, $\mathcal{I}_{\text{HNF}}$, $\mathcal{I}_{\text{PS}}$, and $\mathcal{I}_{\text{HNS}}$, respectively.
As shown in Figure~\ref{fig:intra_contrastive}(b), the proposed intra-sample contrastive loss is applied as follows:

\begin{align}
\begin{split}
    \mathcal{L}_{\text{intra}} = \ & \displaystyle \mathbb{E}_{z\sim \mathcal{I}_{\text{PF}}, z^{+}\sim \mathcal{I}_{\text{PS}}, z^-\sim \mathcal{I}_{\text{HNF}}} \ell(z,z^+,z^-) \\
    + \ & \displaystyle \mathbb{E}_{z\sim \mathcal{I}_{\text{PS}}, z^{+}\sim \mathcal{I}_{\text{PF}}, z^-\sim \mathcal{I}_{\text{\text{HNS}}}} \ell(z,z^+,z^-)
\end{split}
\end{align}
where $\ell$ follows the same contrastive equation as Eq.~\ref{eq:contrastive}.

\vspace{-4mm}
\paragraph{Overall Loss.}
The final loss is the combination of the above three losses,
\begin{equation}
    \mathcal{L}=\mathcal{L}_{\text{cls}} + \beta \cdot \mathcal{L}_{\text{inter}} + \lambda \cdot \mathcal{L}_{\text{intra}}\\
\end{equation}
where $\beta$ and $\lambda$ are hyper-parameters controlling the trade-off between three loss components.

\section{Experiments}

\subsection{Datasets and Implementation Details}

\noindent\textbf{Datasets.}
We evaluate \system on two standard video summarization datasets (SumMe~\cite{gygli2014creating} and TVSum~\cite{song2015tvsum}), two multimodal summarization datasets (Daily Mail~\cite{fu2021mm} and CNN~\cite{fu2021mm}), and a new Behance LiveStream Summarization (BLiSS) dataset.
\textbf{TVSum} dataset consists of 50 videos pertaining to 10 categories. \textbf{SumMe} dataset consists of 25 videos capturing multiple events. 
\textbf{Daily Mail} dataset contains 1,970 samples and \textbf{CNN} dataset contains 203 samples, which are crawled from the news website including video, images, text articles, and captions. We follow the same data split as~\cite{fu2021mm}.
The \textbf{BLiSS} dataset consists of 13,303 pairs of livestream videos and transcribed text, with annotated summaries for both modalities. 

\vspace{0.05in} 
\noindent\textbf{Evaluation Metrics.}
For SumMe and TVSum datasets, following previous work~\cite{zhang2016video,rochan2018video,jung2019discriminative,zhu2020dsnet,narasimhan2021clip,jiang2022joint},  we evaluate the video summarization dataset by the F1 score metric. 
However, as pointed out by~\cite{otani2019rethinking}, the performance of F1 evaluation is mostly determined by the pre-processing segmentation step, and a random method is able to reach similar performance scores. As a result, they propose to utilize rank order statics (Kendall's $\tau$~\cite{kendall1945treatment} and Spearman's $\rho$~\cite{zwillinger1999crc}) as alternative evaluation metrics.
For multimodal summarization datasets, we evaluate the generated text summary by ROUGE~\cite{lin2004rouge} following previous works~\cite{fu2021mm,chen2018iterative,see2017get,fu2020multi}. Specifically, R-1, R-2, and R-L represent ROUGE-1, ROUGE-2, and ROUGE-L F1 scores, respectively, which are widely used to calculate the n-grams overlapping between the output text summary and ground truth text summary.
Same as~\cite{fu2021mm,li2020vmsmo,fu2020multi}, the cosine image similarity is measured between the features of the predicted video summary and ground-truth video summary.

\vspace{0.05in} 
\noindent\textbf{Implementation Details.}
For standard video summarization datasets (SumMe and TVSum), we follow previous work~\cite{gygli2015video,zhang2016video,zhu2020dsnet,jiang2022joint} and use the pre-extracted GoogLeNet~\cite{szegedy2015going} feature as the video input.
To collect the corresponding text modality, we adopt the pre-trained image caption model GPT-2~\cite{radford2019language}\footnote{https://huggingface.co/nlpconnect/vit-gpt2-image-captioning} to generate the caption for each frame.
Next, for all the text modality input, we apply the pre-trained RoBERTa~\cite{liu2019roberta}\footnote{https://huggingface.co/distilroberta-base} to extract textual features for each sentence.
For multimodal summarization datasets (Daily Mail and CNN), we use the same feature extractor as~\cite{fu2021mm,fu2020multi}.
For the BLiSS dataset, pre-trained CLIP~\cite{radford2021learning} and RoBERTa~\cite{liu2019roberta} are adopted to extract features for each frame and each sentence.
The focal loss \cite{lin2017focal} with $\alpha = 0.25$ and $\gamma = 2.0$ is adopted for the classification loss. 
More dataset-specific training/testing details and hyper-parameter choosing are described in the supplementary material.

\begin{table*}[t]
\vspace{-0.2in}
\centering
\caption{Comparison with state-of-the-art methods on the CNN~\cite{fu2021mm} and and Daily Mail~\cite{fu2021mm} datasets.}
\vspace{-0.05in}
\resizebox{.8\textwidth}{!}{
\small
\renewcommand{\arraystretch}{1.15}
    \begin{tabular*}{0.8\linewidth}{@{\extracolsep{\fill}\;}l|l|ccccccc} 
    \toprule
    \multirow{2}{*}{Category} & \multirow{2}*{Method} & \multicolumn{3}{c}{CNN} & \multicolumn{4}{c}{Daily Mail}   \\ 
    \cmidrule(r){3-5}\cmidrule(r){6-9}
    & & R-1 &R-2 &R-L &R-1  &R-2  &R-L &Cos(\%)   \\
    \midrule
    \multirow{2}{*}{Video}
    & VSUMM~\cite{de2011vsumm} &-- &-- &-- &-- &-- &-- &68.74\\
    & DR-DSN~\cite{zhou2018deep} &-- &-- &-- &-- &-- &-- &68.69\\
    & CLIP-It~\cite{narasimhan2021clip} &-- &-- &-- &-- &-- &-- &69.25\\
    \midrule
    \multirow{2}{*}{Text}
    & Lead3~\cite{nallapati2016abstractive} &-- &-- &-- &41.07 &17.87 &30.90 &--\\
    & SummaRuNNer~\cite{nallapati2017summarunner} &-- &-- &--  &41.12 &17.92 &30.94 &--\\
    & NN-SE~\cite{cheng2016neural}   &-- &-- &--  &41.22 &18.15 &31.22 &--\\
    \midrule
    \multirow{6}{*}{Multimodal} & 
    MM-ATG~\cite{zhu2018msmo} &26.83 &8.11 &18.34 &35.38  &14.79  &25.41 &69.17    \\
    & Img+Trans~\cite{hori2019end}   &27.04 &8.29 &18.54 &39.28  &16.64  &28.53  &--   \\
    & TFN~\cite{zadeh2017tensor}  &27.68 &8.69 &18.71 &39.37  &16.38  &28.09 &--       \\  
    & HNNattTI~\cite{chen2018abstractive} &27.61 &8.74 &18.64 &39.58  &16.71 &29.04 &68.76    \\ 
    & $\rm M^{2}$SM \cite{fu2021mm} &27.81  &8.87  &18.73 &41.73 &18.59 &31.68 &69.22     \\ 
    \midrule
    \multirow{3}{*}{Ours} & 
      Video-only & -- & -- & -- & -- & -- & -- & 69.30 \\
    & Text-only & 29.39 & 10.85 & 26.11 & 42.77 & 19.19 & 34.60 & -- \\
    & \bf \system & \bf 30.82 & \bf 11.40 & \bf 27.40 & \bf 44.11 & \bf 20.31 & \bf 35.92 & \bf 70.20 \\
    \bottomrule
    \end{tabular*}
}
\vspace{-0.1in}
\label{tab:cnn_daily}
\end{table*}

\subsection{BLiSS Dataset}
Behance\footnote{http://behance.net/} is a public website with a large amount of livestream videos created by artists showing their work process. The videos are generally hours long and accompanied with transcripts of streamers' speeches. We follow previous work StreamHover~\cite{cho2021streamhover} to expand their dataset to a much larger scale with more modalities.

\vspace{0.05in} 
\noindent\textbf{Data Collection.}
We collected 628 livestream videos with transcripts and other metadata. Each video was further divided into 5-minute long clips for human annotation. Annotators were instructed to select the key sentences from transcripts, and write the text summary and key phrases in their own words for the entire clip.
For each video, we also obtained its thumbnail animation from the website, and selected the most similar frame to the thumbnail from each clip as ground truth key-frames.
More details about the collection process are elaborated in the supplementary material.

\vspace{0.05in} 
\noindent\textbf{Comparison with Existing Multimodal Datasets.}
The BLiSS dataset is much larger than the standard video summarization datasets (SumMe and TVSum) and multimodal summarization datasets (Daily Mail and CNN).
BLiSS has 13,303 data samples and 1,109 total video hours, which is much longer than TVSum (3.5 hours) and Daily Mail (44.2 hours).
For the text modality, the total number of text tokens is 5.4M (BLiSS), greater than 1.3M (Daily Mail) and 0.2M (CNN).
There are other multimodal summarization datasets for the abstractive text summarization task with additional image or video modalities. 
For example, MSMO~\cite{zhu2018msmo}, MMSS~\cite{li2018multi}, VMSMO~\cite{li2020vmsmo} and How2~\cite{sanabria2018how2}. However, none of them have aligned video and text modalities.
Furthermore, we keep some metadata, including the title, streamer information, and audio modality for further potential research.

\begin{table}[t]
\centering
\caption{Comparison with state-of-the-art methods on the SumMe~\cite{gygli2014creating} and TVSum~\cite{song2015tvsum} datasets with F1 scores, Kendall's $\tau$~\cite{kendall1945treatment} and Spearman's $\rho$~\cite{zwillinger1999crc} metrics. We include the results of methods using GoogleNet~\cite{szegedy2015going} features for a fair comparison. \textbf{Bold} and \underline{underline} represent the top-1 and top-2 results.}
\resizebox{\linewidth}{!}{
\renewcommand{\arraystretch}{1.3}
    \begin{tabular}{@{}lcccccc@{}}
    \toprule
    \multirow{2}{*}{Method} & \multicolumn{3}{c}{SumMe} & \multicolumn{3}{c}{TVSum} \\ 
    \cmidrule(l{2pt}r{2pt}){2-4} \cmidrule(l{2pt}r{2pt}){5-7}
        & F1 & $\tau$ & $\rho$ & F1 & $\tau$ & $\rho$ \\ 
        \midrule
        Random~\cite{otani2019rethinking} & 41.0 & 0.000 & 0.000 & 57.0 & 0.000 & 0.000 \\
        Human~\cite{otani2019rethinking} & 54.0 & 0.205 & 0.213 & 54.0 & 0.177 & 0.204 \\
        \midrule
        DR-DSN~\cite{zhou2018deep} & 42.1 & -- & -- & 58.1 &  0.020 & 0.026 \\
        HSA-RNN~\cite{zhao2018hsa} & 42.5 & 0.064 & 0.066 & 44.1 & 0.082 & 0.088 \\
        CSNet~\cite{jung2019discriminative} & 48.6 & -- & -- & 58.5 & 0.025 & 0.034 \\
        VASNet~\cite{fajtl2019summarizing} & 49.7 & -- & -- & 61.4 & -- & -- \\
        DSNet-AB\cite{zhu2020dsnet} & 50.2 & 0.051 & 0.059 & 62.1 & 0.108 & 0.129 \\
        DSNet-AF\cite{zhu2020dsnet} & 51.2 & 0.037 & 0.046 & 61.9 & 0.113 & 0.138 \\
        RSGN~\cite{zhao2021reconstructive} & 45.0 & 0.083 & 0.085 & 60.1 & 0.083 & 0.090 \\
        CLIP-It~\cite{narasimhan2021clip} & 51.6 & -- & -- & \bf 64.2 & 0.108 & 0.147 \\
        iPTNet~\cite{jiang2022joint} & \underline{54.5} & \underline{0.101} & \underline{0.119} & 63.4 & \underline{0.134} & \underline{0.163} \\
        \midrule
        \textbf{\system} & \bf 55.0 & \bf 0.108 & \bf 0.129 & \underline{63.4} & \bf 0.137 & \bf 0.165 \\
        \bottomrule
    \end{tabular}
}
\label{tab:summe_tvsum}
\vspace{-0.2in}
\end{table}

\subsection{Results}
\noindent\textbf{SumMe and TVSum Datasets.}
We compare the proposed method \system with the previous state-of-the-art (SOTA) methods on SumMe \cite{gygli2014creating} and TVSum \cite{song2015tvsum} datasets in Table~\ref{tab:summe_tvsum}.
We first observe that \system achieves the best performance on both datasets. 
Except for the F1 score metric, our \system is slightly worse than CLIP-It~\cite{narasimhan2021clip} but still higher than it for the other two metrics on the TVSum dataset.
CLIP-It also adopts transformer architecture to fuse different modalities by cross-attention, which takes in the generated video caption as text modality. However, it ignores the time correspondence between video and text modalities. 
Instead, our \system aligns cross-modality information and exploits the intrinsic correlation between the video and text at different granularities by our inter-sample and intra-sample contrastive losses.
In addition, the state-of-the-art method iPTNet~\cite{jiang2022joint} utilizes an additional moment localization dataset Charades-STA~\cite{gao2017tall} to help address the data scarcity problem but results in a much longer training time, however, without utilizing extra datasets, our \system can still outperform it on all the metrics, which strongly justifies the superiority of our design.

\vspace{0.05in} 
\noindent\textbf{Daily Mail and CNN Datasets.}
As shown in Table~\ref{tab:cnn_daily}, we also compare our \system with previous methods on the CNN \cite{fu2021mm} and Daily Mail \cite{fu2021mm} datasets. 
Since the text modality of CNN and Daily Mail datasets do not have time information, we only apply the inter-sample and intra-sample contrastive losses without the alignment-guided self-attention module.
We first observe that \system can indeed greatly benefit from leveraging the multimodal information, which boosts the text summary metric by 1-2\% ROUGE F1 score and increases the video summary cosine similarity by 0.9\%. 
Compared to the state-of-the-art multimodal method $\rm M^{2}$SM~\cite{fu2021mm}, which utilizes additional transcript extracted from videos as the bridge between video and text modality,
\system is better by 3\% and 2.4\% in ROUGE-1 F1 score on two datasets respectively.
For the video summarization, our transformer-based \system can outperform multimodal summarization method $\rm M^{2}$SM~\cite{fu2021mm} and state-of-the-art video summarization model CLIP-It~\cite{zhou2018deep} by 1\%.

\begin{table*}[t]
\vspace{-0.2in}
\centering
\small
\begin{minipage}{.48\textwidth}
    \centering
    \caption{Comparison results on the BLiSS dataset. Note that BART \cite{lewis2019bart} is abstractive summarization based method, and the rest are extractive summarization based. }
    \vspace{-0.1in}
    \resizebox{\linewidth}{!}{
    \renewcommand{\arraystretch}{1.15}
    \begin{tabular}{@{}l|ccccccc|c@{}} 
        \toprule
        Method & R-1  & R-2  & R-L &Cos(\%)   \\
        \midrule
        DSNet-AF~\cite{zhu2020dsnet} & -- & -- & -- & 62.70 \\
        CLIP-It~\cite{narasimhan2021clip} & -- & -- & -- & 63.58 \\
        Miller~\cite{miller2019leveraging} & 40.90 & 26.48 & 39.14 & -- \\
        BART~\cite{lewis2019bart} & 49.11 & 38.59 & 48.08 & -- \\
        SummaRuNNer~\cite{nallapati2017summarunner} & 49.70 & 38.00 & 48.51 & --\\
        \midrule
        \bf \system  & \bf 52.61 & \bf 41.88 & \bf 51.52 & \bf 64.61 \\
        \bottomrule
    \end{tabular}
    }
    \label{tab:behance}
\end{minipage}
\hfill
\begin{minipage}{.5\textwidth}
    \centering
    \caption{Contribution of each component on the SumMe dataset. ``Align.'', ``Inter.'', and ``Intra.'' represent the alignment-guided self-attention, inter-sample and intra-sample contrastive loss, respectively.}
    \vspace{-0.1in}
    \resizebox{\linewidth}{!}{
    \renewcommand{\arraystretch}{1.15}
    \begin{tabular}{@{}l|ccc|ccc@{}}
        \toprule
        Inputs & Align. & Inter. & Intra. & F1 & $\tau$ & $\rho$  \\
        \midrule
        Video-Only &  &  &  & 49.8 & 0.070 & 0.084 \\ 
        \midrule
        \multirow{5}{*}{Multimodal} 
        &  &  &  & 50.5 & 0.083 & 0.096   \\
        & \checkmark &  &  & 51.5 & 0.089 & 0.104   \\
        & \checkmark & \checkmark &  & 52.5 & 0.095 & 0.110   \\
        & \checkmark &  & \checkmark & 54.0 & 0.102 & 0.121   \\
        & \checkmark & \checkmark & \checkmark & \bf 55.0 & \bf 0.108 & \bf 0.129 \\
        \bottomrule
    \end{tabular}
    }
    \label{tab:ablation}
\end{minipage}
\end{table*}

\begin{figure*}[t]
\vspace{-0.15in}
\centering
    \adjincludegraphics[width=\linewidth, trim={{0.04\width} {0.27\height} {0.05\width} {0.24\height}},clip]{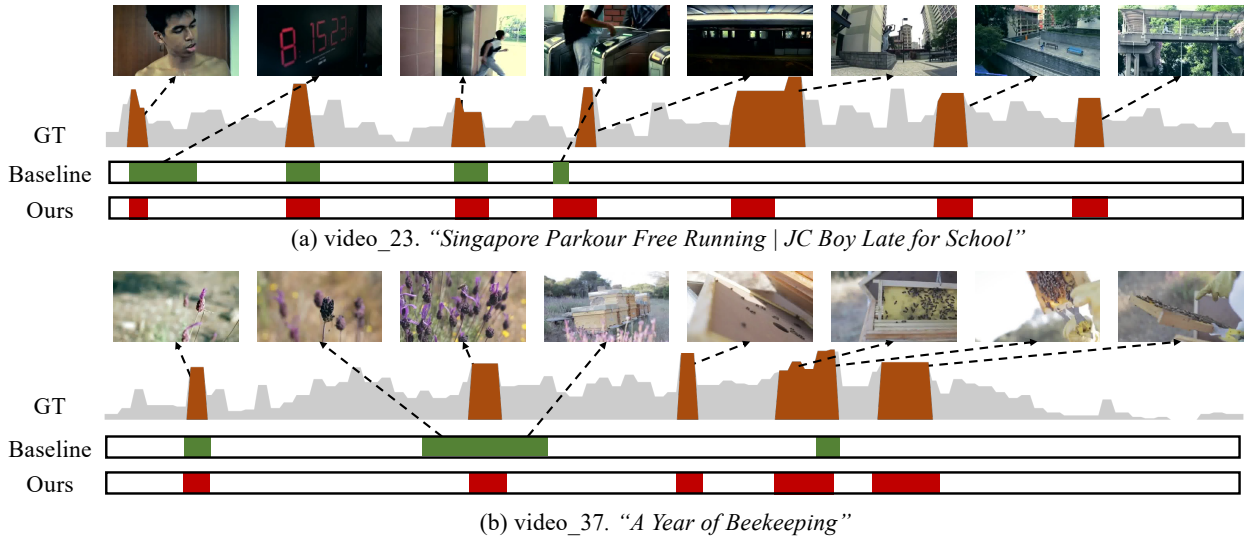}
    \vspace{-0.25in}
    \caption{Example summarization results for the TVSum dataset. Titles are shown for each video. ``Baseline'' denotes our \system without the alignment module and dual contrastive losses. The gray histogram shows the ground-truth importance scores for each frame.}
\vspace{-0.18in}
\label{fig:visualization}
\end{figure*}

\vspace{0.05in} \noindent\textbf{BLiSS Dataset.}
We validate \system on the livestream videos from the BLiSS dataset by comparing it with existing video and text summarization methods.
As shown in Table~\ref{tab:behance}, when comparing with video summarization methods DSNet-AF~\cite{zhu2020dsnet} and CLIP-It~\cite{narasimhan2021clip}, our \system achieves the best results on the video cosine similarity metric. 
Although CLIP-It also utilizes the additional text modalities by the cross-attention operation, \system can still outperform it by 1\%.
Compared to the extractive text summarization method SummaRuNNer~\cite{nallapati2017summarunner} and the abstractive text summarization method BART~\cite{lewis2019bart}, \system outperforms both of them by at least 3\% on all the ROUGE scores.
It further demonstrates the superior effectiveness of \system on livestream videos.

\subsection{Ablation Studies}
To further investigate the contribution of each component in \system, we conduct ablation studies in Table~\ref{tab:ablation}.
We first observe that adding the text modality input can enhance the final results of video summarization.
However, as we mentioned before, without alignment of video and text modalities, directly applying global attention between untrimmed video and text input tends to introduce too much noise and result in inferior performance.
After we align and attend the video and text modalities with the proposed alignment-guided self-attention module, we can improve the F1 score by 1\%.
Furthermore, for the dual contrastive losses, it is obvious that a consistent gain can be achieved by adding either one of these two losses. 
In particular, introducing the intra-sample contrastive loss significantly increases the performance by 2.5\%.
It proves that exploring the intrinsic mutual correlation between video and text and mining hard negative samples can greatly enhance the ability to localize the important frames and sentences.
In addition, two contrastive losses are complementary to each other.
When incorporating all three proposed components together, our approach boosts the final performance from 50.5\% to 55.0\%.

\subsection{Visualization}
Figure~\ref{fig:visualization} shows the visual comparison between baseline and \system.
We observe that the typical errors of the baseline model can be addressed by the proposed alignment module and dual contrastive losses, such as missing detection of important segments and inaccurate summary boundary prediction.
It further verifies the effectiveness of \system.
More visualizations on the BLiSS dataset are provided in the supplementary material.

\vspace{-0.05in}
\section{Conclusion}
\vspace{-0.05in}
In this paper, we present \system, a novel unified transformer-based framework for multimodal summarization. \system is designed to align and attend different modalities by leveraging time correspondences that previous methods neglect.
Also, we introduce dual contrastive losses to exploit the inter-sample and intra-sample cross-modality information.
Extensive experiments on multiple datasets validate the effectiveness of our \system.
In addition, we collect a large-scale multimodal summarization dataset focusing on livestream videos and transcripts. We hope it can be beneficial for further research in this area.

\medskip
\noindent\textbf{Acknowledgements.} This work was partially supported by DARPA SemaFor (HR001119S0085) program and gifts from Adobe.

\clearpage
{\small
\bibliography{egbib,main}
}
\clearpage

\appendix
\section*{Appendix}

Sec.~\ref{sec:bliss} elaborates more details about the collection process of BLiSS dataset.
Sec.~\ref{sec:experiment_details} provides more dataset-specific implementation details and hyper-parameters for training and testing.
We also present more qualitative results in Sec.~\ref{sec:qualitative_more}.
Finally, we discuss the limitation and some future work of our paper in Sec.~\ref{sec:limitation}.

\section{BLiSS Dataset}
\label{sec:bliss}

\noindent\textbf{Data Collection}

We collected 628 livestream videos from behance.net, including the corresponding English transcripts and meta data.
Audio tracks are also available in the videos; however, we don't use them in our study since most information from audio modality can be captured by transcripts. Meta data from the website are annotated by creators which include title, overall text description, creative fields, creative tools, streamer, cover image and animation.

The transcripts are first segmented by sentence, each with corresponding time stamps. We remove transcripts with very short duration which are likely caused by broken words and speech recognition failures. Based on the transcript segments, we further divide each video into 5-minute long clips, so that each clip has its corresponding frames and aligned transcript sentences.

Annotators were instructed to watch the entire clip, read the transcript sentences, and select keywords from the sentences representing the important content in the clip. Each clip has about 5 to 10 keywords. The sentences containing keywords are regarded as key sentences for extractive text summarization. The annotators were also asked to write a summary of the whole clip in their own words, which can be used for abstractive text summarization.

For each video, we extract all the frames from its thumbnail animation. 
For each frame $g$ in the thumbnail, we select the most similar frame $f$ in the video as the key-frame.

\vspace{0.05in} 
\noindent\textbf{Corpus Statistics of BLiSS Dataset}
In Table~\ref{tab:statistic}, we compare the statistics of our collected BLiSS dataset with other datasets including standard video summarization datasets (SumMe~\cite{gygli2014creating} and TVSum~\cite{song2015tvsum}), multimodal datasets (CNN~\cite{fu2021mm} and Daily Mail~\cite{fu2021mm}) and the transcript summarization dataset (~\cite{cho2021streamhover}).
We can see that our BLiSS dataset has a much larger scale than all the other datasets. 
Specifically, the BLiSS dataset has 1,109 hours of total video duration. 
The total number of text tokens of the BLiSS dataset is 5.5M, much larger than the Daily Mail and StreamHover datasets.

\begin{table*}[t]
\centering
\caption{Statistics comparison of BLiSS dataset with other datasets.}
\resizebox{0.85\linewidth}{!}{
\renewcommand{\arraystretch}{1.2}
    \begin{tabular*}{0.85\linewidth}{@{\extracolsep{\fill}\;}l|cccccc} 
    \toprule
     & SumMe & TVSum & CNN & Daily Mail & StreamHover & BLiSS  \\
    \midrule
    Number of Data &  25 & 50 & 203 & 1970 & 5421 & 13303 \\
    Total Video Duration (Hours) & 1.0 & 3.5 & 7.1 & 44.2 & 452 & 1109 \\
    Total Number of Text Tokens & -- & -- & 0.2M & 1.3M & 3.1M & 5.5M \\
    Avg. Video Summary Length & 44 & 70 & -- & 2.9 & -- & 10.1 \\
    Avg. Text Summary Length & -- & -- & 29.7 & 59.6 & 79 & 49 \\
    \bottomrule
    \end{tabular*}
}
\label{tab:statistic}
\end{table*}

\begin{figure*}[t]
\vspace{-0.1in}
\centering
    \adjincludegraphics[width=\linewidth, trim={{0.03\width} {0.15\height} {0.03\width} {0.15\height}},clip]{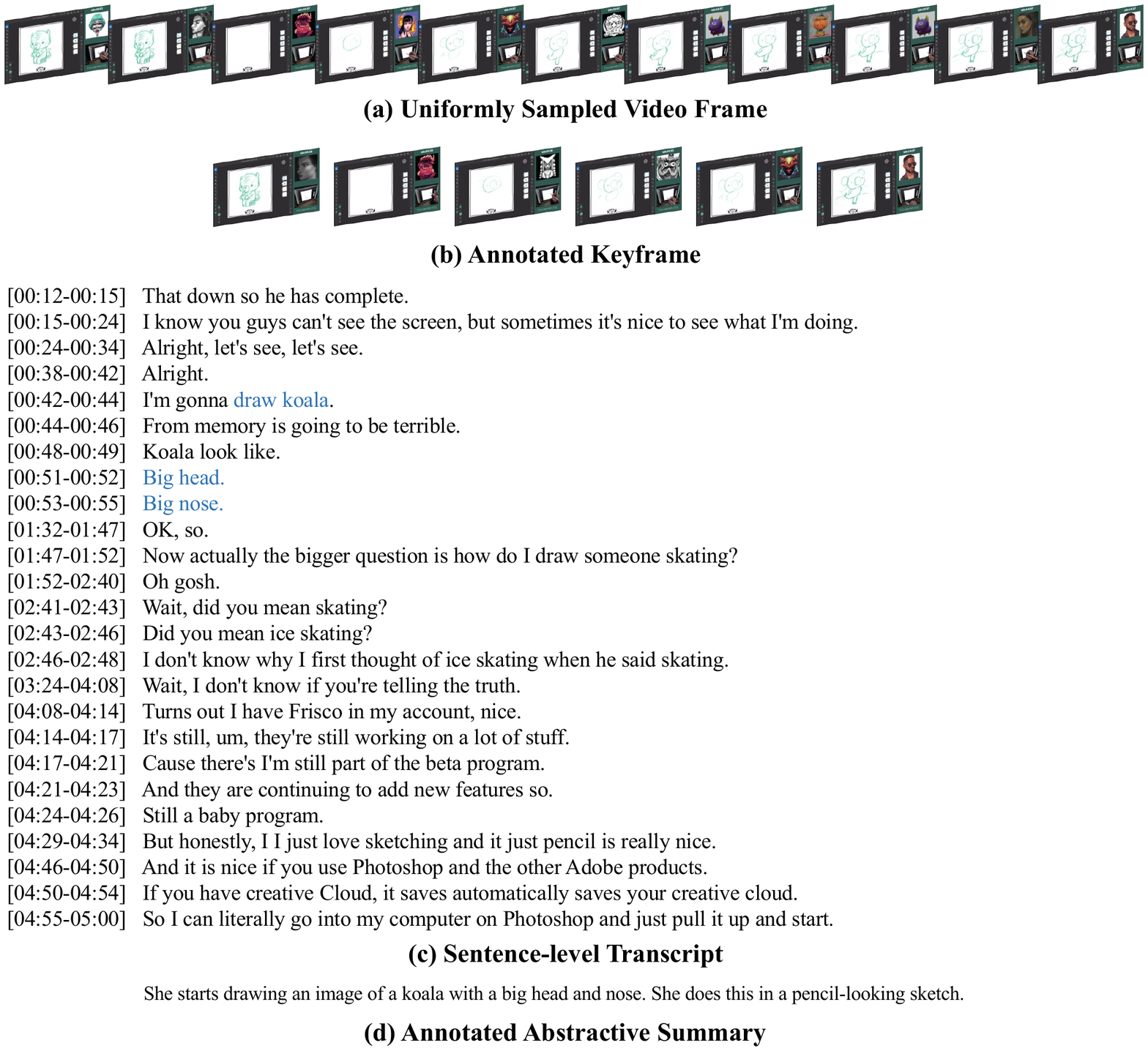}
    \vspace{-0.25in}
    \caption{Example of one data sample from the BLiSS dataset. Here, we visualize the uniformly sampled video frames, annotated keyframes, sentence-level transcript, and abstractive text summary. Note that the extractive text summary is formed by the key sentences, where the ground-truth keywords are marked with \textcolor{-red!70!green}{blue} color. Best viewed in color.}
\vspace{-0.15in}
\label{fig:example_bliss}
\end{figure*}

\vspace{0.05in} 
\noindent\textbf{Example}
We show one example of the annotated sample in the BLiSS dataset in Figure~\ref{fig:example_bliss}. We visualize the uniformly sampled video frames, annotated keyframes, sentence-level transcripts, and the abstractive text summary. Note that the extractive text summary is formed by the key sentences, where the ground-truth keywords in the key sentences are marked in blue color.

\section{Experiment Details}
\vspace{-0.05in}
\label{sec:experiment_details}
On multimodal summarization datasets (Daily Mail and CNN), we train our \system with a batch size of 4, a learning rate of 2e-4, weight decay of 1e-7 and 1e-5, training epochs of 100,  $L$ = 2, the ratio controlling hard-negative samples $r=8$, the balancing weights for dual contrastive losses $\beta$ of 0.001 and 0, $\lambda$ of 0.001 and 0 for the Daily Mail and CNN datasets, respectively. 

On standard video summarization datasets (SumMe and TVSum), we train our \system with a batch size of 4, a learning rate of 1e-3, weight decay of 1e-3 and 1e-5, training epochs of 300, number of transformer layers $L$ = 2, the ratio controlling hard-negative samples $r=16$ the balancing weights for dual contrastive losses $\beta$ of 0.1, $\lambda$ of 3 and 1 for the SumMe and TVSum datasets, respectively.

On the BLiSS dataset, we set a batch size of 64, a learning rate of 1e-3, weight decay of 1e-7, training epochs of 50, transformer layers $L$ of 6, the ratio controlling hard-negative samples $r=4$, the balancing weights for dual contrastive losses $\beta$ of 0.01 and $\lambda$ of 0.001.

We set the expansion size for both sides of key-frames and key-sentences in the contrastive pair selection procedure as 4 on all the datasets.

\begin{figure*}[t]
\vspace{-0.1in}
\centering
    \adjincludegraphics[width=\linewidth, trim={{0.03\width} {0.12\height} {0.065\width} {0.07\height}},clip]{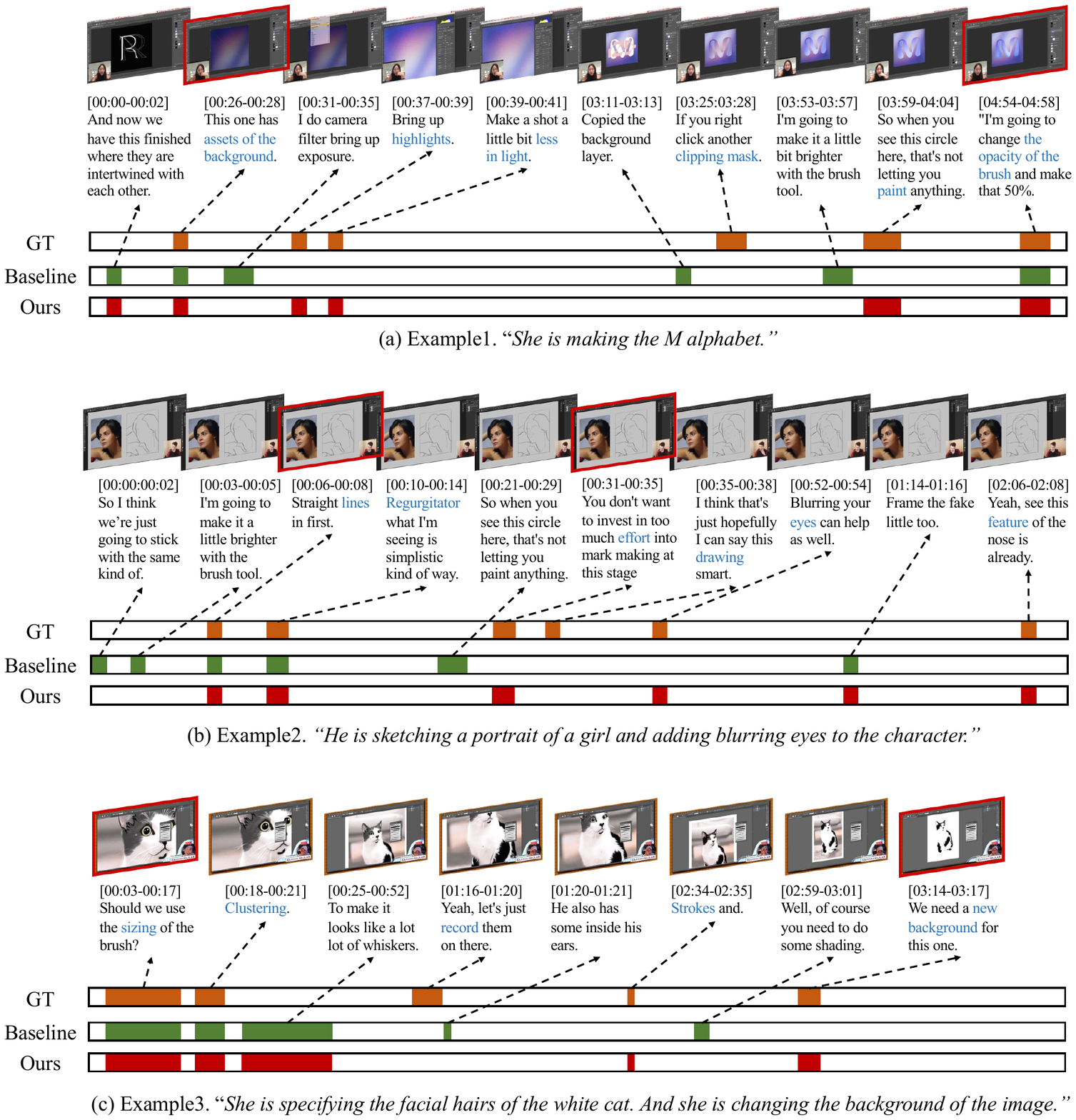}
    \caption{Visualization of multimodal summarization results for the BLiSS dataset. The ground-truth text summary, predictions from the baseline model and our \system are shown for each video. ``Baseline'' denotes our \system without the proposed alignment module and dual contrastive losses. The ground-truth keywords from key sentences are marked with \textcolor{-red!70!green}{blue} color. We also show the corresponding video frames for each transcribed sentence where the frames with \textcolor{red}{red} boxes represent some of the predicted key-frames from our \system. The title for each video clip is the annotated abstractive summary. }
\vspace{-0.15in}
\label{fig:visualization_behance}
\end{figure*}

\section{More Qualitative Results}
\label{sec:qualitative_more}
\vspace{-0.05in}

In Figure~\ref{fig:visualization_behance}, we show three different examples of multimodal summarization results on the BLiSS dataset.
We can see that, compared to the baseline method, our \system can predict the key sentences more accurately and faithfully for the extractive text summarization task. 
It proves the effectiveness of the proposed alignment module and dual contrastive losses for the text modality.
For the video summarization task, because livestream videos change slowly over time, their video frames generally share similar visual content. However, our predicted key-frames can still capture the important scenes from the input video qualitatively.

\section{Limitation and Future Work}
\label{sec:limitation}
\vspace{-0.05in}

The main limitation is that our \system is based on the Transformer~\cite{vaswani2017attention} architecture with the self-attention operation, which suffers from heavy computation cost due to the quadratic computation complexity with respect to the input sequence length.
Although there are a series of works~\cite{beltagy2020longformer,kitaev2020reformer,correia2019adaptively,vyas2020fast} trying to design computation efficient transformer models to handle long sequences, it is out of the scope of our paper and we still follow the basic transformer design.
In addition, the data annotation process for the video and text summaries is laborious. More research on unsupervised or self-supervised multimodal summarization tasks would be a good direction for future work.

\end{document}